\documentclass[10pt,twocolumn,letterpaper]{article}

\usepackage{wacv}              
\usepackage[accsupp]{axessibility}
\usepackage{multirow}
\usepackage{float}
\usepackage{dblfloatfix}  
\usepackage{subcaption}
\usepackage{balance}
\usepackage{soul}
\usepackage{tabularx}
\usepackage{booktabs}
\usepackage{multirow}
\usepackage{array}

\definecolor{wacvblue}{rgb}{0.21,0.49,0.74}
\usepackage[pagebackref,breaklinks,colorlinks,allcolors=wacvblue]{hyperref}

\usepackage{comment}

\def\wacvPaperID{9} 
\def\confName{WACV}
\def\confYear{2026}

\title{Neighborhood Feature Pooling for Remote Sensing Image Classification}

\author{First Author\\
Institution1\\
Institution1 address\\
{\tt\small firstauthor@i1.org}
\and
Second Author\\
Institution2\\
First line of institution2 address\\
{\tt\small secondauthor@i2.org}
}
\author{%
  \begin{tabular}{@{}c@{}}
    Fahimeh Orvati Nia$^{1}$, Amirmohammad Mohammadi$^{1}$, Salim Al Kharsa$^{1}$,
    Pragati Naikare$^{2}$, \\ Zigfried Hampel–Arias$^{1,3}$,
    and Joshua Peeples$^{1,3}$\\[6pt]
    $^{1}$Dept.\ of Electrical \& Computer Engineering, Texas A\&M University, College Station, TX, USA\\
    $^{2}$Dept.\ of Computer Science \& Engineering, Texas A\&M University, College Station, TX, USA\\
    $^{3}$Los Alamos National Laboratory, Los Alamos, NM, USA\\
  \end{tabular}
}

\begin{document}
\maketitle
\begin{abstract}
In this work, we propose neighborhood feature pooling (NFP) as a novel texture feature extraction method for remote sensing image classification. The NFP layer captures relationships between neighboring inputs and efficiently aggregates local similarities across feature dimensions. This new approach is implemented using convolutional layers and can be seamlessly integrated into any network. Results comparing the baseline models and the NFP method indicate the potential of this new approach for classification. NFP consistently improves performance across diverse datasets and architectures while maintaining minimal parameter overhead. The code for this work is publicly available\footnote{\url{https://github.com/Advanced-Vision-and-Learning-Lab/Neighbour_Feature_Pooling}}.
\end{abstract}
    
\section{Introduction}
\label{sec:intro}

Remote sensing has various applications in environmental monitoring~\cite{Kerr2003}, urban planning~\cite{Wellmann2020Urban}, and agriculture~\cite{Mulla2013}. Machine learning, particularly convolutional neural networks (CNNs), has been used in processing and analyzing remote sensing images~\cite{Zhang2016Tutorial,Li2018Survey}. In high-resolution remote sensing imagery, fine-grained textural patterns can often distinguish different land-cover or scene types~\cite{puissant2005utility}. For example, in the University of California Merced (UCMerced) Land Use dataset~\cite{Yang2010UCMerced} and the Northwestern Polytechnical University-Remote Sensing Image Scene Classification (NWPU-RESISC45) benchmark~\cite{Cheng2017RESISC45}, visual distinctions between categories often lie in recurrent textures or spatial arrangements (\eg, the grid-like layout of residential blocks versus the irregular granular pattern of forest canopies) rather than unique objects~\cite{Cheng2017RESISC45}.

However, standard CNN architectures can inadvertently lose these cues when using traditional pooling operations (e.g., average and max pooling) \cite{peeples2021histogram}. These layers are effective in summarizing information at reduced spatial dimensions, yet typically fail to encode important structural details~\cite{zhao2024improved}. This limitation stems from the lack of explicit weighting on neighboring relationships within the feature space. To address this gap, neighborhood feature pooling (NFP) is introduced in this work. This is a novel layer that measures how similar each neighbor is to a center pixel and/or feature vector. NFP can work alongside the usual global average pooling (GAP), adding a texture-aware branch for improved feature representation.

\section{Related Work}
\label{sec:related}

\begin{figure*}[htb]
    \centering

    \begin{subfigure}[b]{0.71\linewidth}
        \centering
        \includegraphics[width=\linewidth]{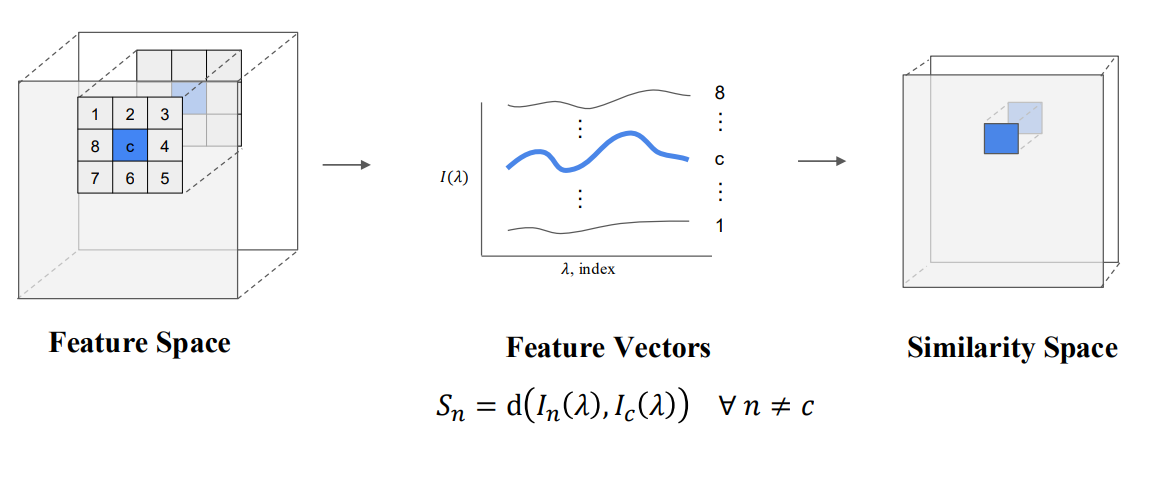}
        \caption{{Overview of Neighborhood Feature Pooling (NFP).}
        For each center pixel \( I_c(\lambda) \), a similarity score is computed with each neighbor \( I_i(\lambda) \) using a function \( d(\cdot,\cdot) \). These values form the channels of a similarity feature vector.}
    \end{subfigure}

    \vspace{0.5em}

    \begin{subfigure}[b]{0.71\linewidth}
        \centering
        \includegraphics[width=\linewidth]{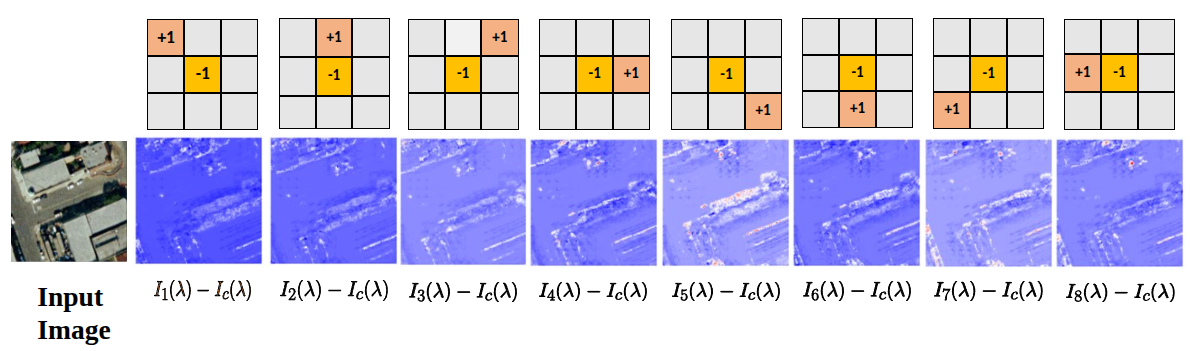}
        \caption{An input patch from the UC Merced dataset (``buildings'' class) is processed with eight directional difference kernels, which measure relative feature differences between the center pixel and its neighbors (scaled dot product similarity). 
For visualization, the outputs are normalized to the range [0,1] and computed with a dilation factor of 15, yielding NFP maps that form a multi-channel similarity representation of local texture relationships, later aggregated into the NFP feature vector.}

    \end{subfigure}

\caption{Overview of the NFP framework.
(a) NFP example of feature aggregation process.
(b) Example with \(r=1\), showing the input patch, directional kernels, and the resulting similarity maps that encode local texture relationships.
Together, these demonstrate how NFP captures local structural information that is later integrated with global features (\cref{fig:architecture}).}

    \label{fig:example}
\end{figure*}

Several methods have been proposed to address the challenge of encoding structural detail for texture recognition. For example, Local Binary Patterns (LBP)~\cite{ojala1994performance} capture texture information by encoding the difference in the center pixel between each of its neighbors and computing a weighted sum based on the binary position of the neighboring pixel. However, LBP operates on grayscale intensities and discards magnitude information~\cite{ahmed2011compound}, limiting its ability to encode fine variations in images. Several advanced texture encoding and pooling methods have been proposed to improve texture representation. Deep Texture Encoding Network (DeepTEN) \cite{zhang2017deep} leveraged dictionary learning to encode texture information in learnable codewords that captured robust features. Randomized Deep Activation Map pooling (RADAM)~\cite{scabini2023radam} samples and aggregates deep activation maps in a randomized manner to capture a wide range of texture patterns via stochastic encoding. Fractal-based pooling techniques, including implementations based on lacunarity~\cite{florindo2024fractal,mohan2024lacunarity}, quantify texture complexity at multiple scales by analyzing the distribution of gaps in local pixel density across spatial patterns.

Texture descriptors in images may be statistical, summarizing pixel-value distributions, or structural, modeling relationships among neighboring pixels~\cite{peeples2021histogram}. Two structural approaches exploit similarity maps to enrich convolutional features. The Local Similarity Pattern (LSP) layer appends cosine-similarity maps between each pixel and its eight-neighbor window, supplying additional structural cues for stereo-matching networks \cite{Liu2022LSP}. In multi-sensor anomalous change detection, a Neighborhood Similarity Feature Space (NSFS) concatenates dot-product similarity responses as extra channels, boosting texture discrimination without resampling \cite{HampelArias2023FeatureLayers}. While these methods demonstrate the benefit of neighbor-wise similarity, LSP and NSFS only used cosine similarity and did not investigate the use of other measures to quantify neighborhood relationships as well as these methods were not implemented using convolutional layers that can be readily added to any network in different locations.

\section{Method}
\label{sec:method}
\begin{figure*}[htb]
  \centering
  \includegraphics[width=0.88\linewidth]{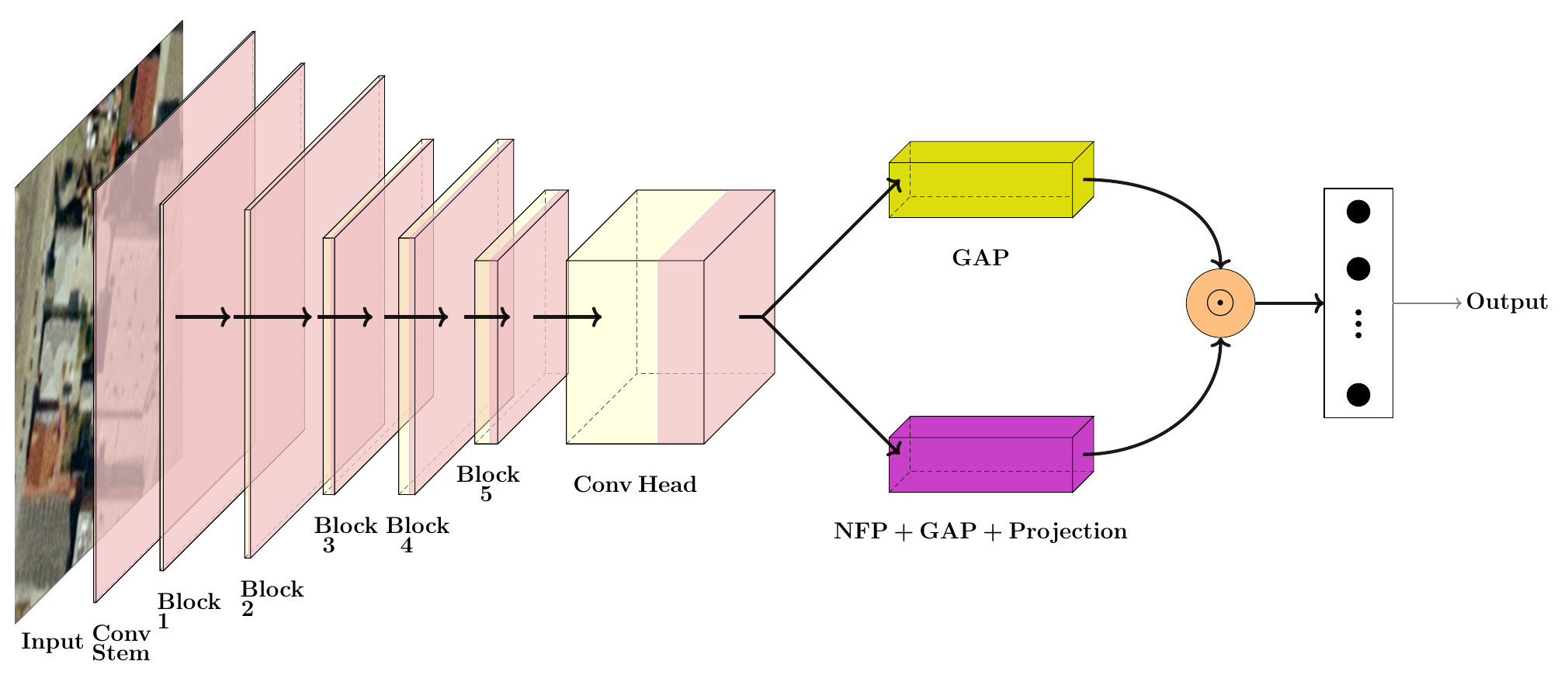}
  \caption{
    Full architecture illustration of the proposed model using MobileNetV3~\cite{howard2019searching} with NFP. Each block represents a stage in the feature extraction pipeline. After feature maps are extracted from the input, the features are aggregated through two branches: global average pooling (GAP) and NFP. The NFP branch first extracts the similarity maps then the similarity values are aggregated through GAP. The average NFP features are then upsampled to the same dimension using a $1 \times 1$ convolution. The final step is for the GAP and NFP feature vectors to be multiplied before being passed into the output classification layer. The $\odot$ symbol denotes element-wise multiplication used to fuse the GAP and NFP feature vectors.}

  \label{fig:architecture}
\end{figure*}

\subsection{Neighborhood Similarity Computation}
\label{subsec:nfp-construction}

The core operation of NFP involves comparing each center pixel with its surrounding neighbors using a feature-space similarity function. Let \( I_n(\lambda) \in \mathbb{R}^d \) denote the feature vector of a pixel at spatial position \( n \), where \( \lambda \in \{1, \dots, d\} \) indexes the feature dimension. For a neighborhood of radius \( r \) centered at position \( c \), the similarity \( S_n \) between the center \( I_c(\lambda) \) and each neighbor \( I_n(\lambda) \) is defined as

\begin{equation}
S_n = d(I_n(\lambda), I_c(\lambda)), \quad \forall n \ne c,
\label{eq:nfp_similarity}
\end{equation}

where \( d(\cdot, \cdot) \) is a similarity function such as cosine or dot product. This operation yields a set of \( (2r+1)^2 - 1 \) scalar values per center pixel, producing similarity feature maps that encode local texture structure, as illustrated in \cref{fig:example}. NFP also employs a predefined similarity function (\textit{e.g.},cosine similarity), but applies it locally within each feature map and aggregates the resulting neighborhood responses as a lightweight layer. Since the backbone and projection layers are trained end to end, the feature embeddings entering the similarity computation adapt during learning, so the similarity patterns themselves evolve even though the metric form is fixed. This localized, similarity-driven pooling preserves structural context while adding little overhead, making NFP a practical alternative to methods that require more parameters or codebooks such as DeepTEN and RADAM in remote-sensing classification.

Similarity measures such as cosine or Pearson correlation can yield negative values, which are retained to represent dissimilar or anti-correlated relationships between neighbors. For distance-based metrics such as \( L_p \) norm, the computed distances are negative so that higher similarity consistently corresponds to larger values (i.e., values closer to zero indicate more similarity). This convention provides a unified interpretation across metrics and ensures that all similarity and distance functions are handled within a common framework. The current NFP layer uses a radius of $r = 1$ to capture strictly local relationships while avoiding the quadratic growth in similarity computations and redundancy with receptive field expansion already provided by deeper convolutional layers which results in 8 similarity feature maps given any input feature dimension $d$. Additionally, this feature is inspired from LBP which defines a radius and number of equally spaced pixels, $P$, to define the neighborhood \cite{ojala2002multiresolution}. LBP generally uses a circular neigbhorhood, but the proposed NFP leverages convolutional layers in Pytorch that use a square neighborhood. To keep our method comparable to the original LBP method, a radius of 1 and $P$ = 8 is used to be similar to a $3 \times 3$ square neighborhood. The formulation in \cref{eq:nfp_similarity} thus provides an explicit mechanism for encoding relative feature-space relationships across local neighborhoods, preserving structural texture cues often discarded by global pooling.

\subsection{Similarity Function Options}
\label{subsec:similarity-measures}

The similarity function \( d(\cdot, \cdot) \) in \cref{eq:nfp_similarity} can be drawn from a wide set of metrics, each emphasizing different aspects of feature space geometry. The choice of metric can influence NFP by defining how local neighborhood relationships are quantified.
 We follow the taxonomy of the first three categories defined by Deborah et al.~\cite{7061924} as the feature vectors are similar to hyperspectral signatures used in remote sensing. We also introduced additional similarity metrics inspired by deep learning methods:

\begin{itemize}
    \item \textbf{Category~1 (Vector in Euclidean space):} L$_p$ norm, RMSE, Geman--McClure, Canberra, dot product (vanilla and scaled), cosine similarity, and sharpened cosine~\cite{wu2023exploring}
    
    \item \textbf{Category~2 ($n$-dimensional data in manifold):} Goodness-of-Fit Coefficient (GFC)

    \item \textbf{Category~3 (Distribution):} Chi-squared (two formulations), Hellinger, Jeffrey divergence, Squared-Chord, Pearson correlation, Smith’s measure, and Earth Mover’s Distance (EMD)~\cite{avi2023differentiable}.
    

\end{itemize}

\noindent This categorization preserves the theoretical grounding of~\cite{7061924} while covering all similarity measures implemented in the NFP module, ensuring both principled analysis and adaptability to modern feature representations. Cosine similarity is selected as the default due to its empirical performance across datasets. However, the modularity of NFP enables swapping any of the similarity functions, as further explored in~\cref{subsec:similarity-analysis}.

\begin{table*}[t]
\centering
\footnotesize
\setlength{\tabcolsep}{4pt}

\begin{tabularx}{\textwidth}{@{}l l *{3}{>{\centering\arraybackslash}X}@{}}
\toprule
\textbf{Backbone} & \textbf{Method} & \textbf{RESISC45} & \textbf{UC Merced} & \textbf{EuroSAT} \\
\midrule

\multirow{6}{*}{ResNet18}
& GAP   & 91.28 ± 0.04 (11.20) & 96.62 ± 0.19 (11.19) & 98.27 ± 0.19 (11.21) \\
& Lacunarity & 93.10 ± 0.01 (11.20) & 97.30 ± 0.11 (11.19) & 98.23 ± 0.33 (11.21) \\
& Fractal    & 92.29 ± 0.16 (11.46) & 97.62 ± 0.34 (11.45) & 98.49 ± 0.20 (11.48) \\
& RADAM      & 92.08 ± 0.15 (11.20) & 97.46 ± 0.22 (11.19) & 98.40 ± 0.23 (11.21) \\
& DeepTEN    & 93.19 ± 0.05 (11.96) & 97.14 ± 0.58 (11.57) & 98.50 ± 0.28 (11.42) \\
& NFP (Ours) & \textbf{93.22 ± 0.25} (11.20) & \textbf{98.86 ± 0.58} (11.19) & \textbf{98.52 ± 0.20} (11.22) \\
\midrule

\multirow{6}{*}{MobileNetV3}
& GAP   & 93.48 ± 0.37 (4.25) & 97.57 ± 0.57 (4.22) & 98.23 ± 0.13 (4.21) \\
& Lacunarity & 92.93 ± 0.41 (4.25) & 97.94 ± 0.59 (4.22) & 98.52 ± 0.32 (4.21) \\
& Fractal    & 93.55 ± 0.19 (5.17) & 97.94 ± 0.22 (5.15) & 98.35 ± 0.15 (5.14) \\
& RADAM      & 85.59 ± 2.61 (4.25) & 97.78 ± 1.81 (4.22) & 98.27 ± 0.05 (4.21) \\
& DeepTEN    & \textbf{94.95 ± 0.09} (5.68) & 98.49 ± 0.31 (4.94) & 98.35 ± 0.23 (4.60) \\
& NFP (Ours) & 94.80 ± 0.21 (4.25) & \textbf{98.49 ± 0.28} (4.23) & \textbf{98.53 ± 0.12} (4.22) \\
\midrule

\multirow{6}{*}{ConvNeXt-Atto}
& GAP   
& 93.90 ± 0.27 (3.39) 
& 96.19 ± 0.78 (3.38) 
& 96.33 ± 0.26 (3.38) \\

& Lacunarity 
& 93.25 ± 0.36 (3.39)
& 94.37 ± 0.66 (3.38) 
& 96.90 ± 0.36 (3.38) \\

& Fractal    
& 90.58 ± 0.56 (3.43) 
& 95.79 ± 0.22 (3.42) 
& 96.38 ± 0.35 (3.42) \\

& RADAM      
& 91.61 ± 0.84 (3.39) 
& 95.63 ± 0.22 (3.38) 
& 97.16 ± 0.26 (3.38) \\

& DeepTEN    
& 94.27 ± 0.31 (3.85) 
& 96.27 ± 0.30 (3.60) 
& \textbf{97.76 ± 0.17} (3.49) \\

& NFP (Ours) 
& \textbf{94.71 ± 0.21} (3.39) 
& \textbf{97.30 ± 1.07} (3.38) 
& 97.65 ± 0.26 (3.38)  \\

\bottomrule

\end{tabularx}

\caption{Classification accuracy (\%) and model size (in millions of parameters) across three remote-sensing datasets. Each cell shows average accuracy $\pm$ 1 standard deviation, followed by parameter count in parentheses. For each backbone architecture, the best performing (i.e., highest average accuracy) method on each dataset is \textbf{bolded}.}
\label{tab:final_grouped_accuracy}
\end{table*}





\subsection{Model Architecture Integration}
\label{subsec:model-arch}

As illustrated in \cref{fig:architecture}, NFP is inserted after the final backbone stage and before the pooling/classification head. This configuration is used for all main results, while alternative placements (early-stage or multi-stage integration) are explored in \cref{subsec:layer-replacement}. Given backbone feature maps \(X \in \mathbb{R}^{B \times C \times H \times W}\), NFP constructs local neighborhoods of radius \(r\), corresponding to a kernel size \(k = 2r+1\) and \(N_r = k^2 - 1\) neighbors. Two depthwise (channel-wise) convolutions with fixed, sparse \(k \times k\) kernels are applied: one selects the center feature, while the other gathers the \(N_r\) neighbors. For distance-based variants, the kernel assigns \(+1\) to the center and \(-1\) to the neighbor. These operations produce an affinity stack \( S \in \mathbb{R}^{B \times N_r \times H' \times W'} \), where \( H' = H - (k-1) \) and \( W' = W - (k-1) \).

The affinity stack is spatially averaged and projected to a vector in \( \mathbb{R}^{C'} \) with \( C' = C \). A \(1 \times 1\) convolution is applied only if \( C' \ne C \). The resulting vector is fused with the backbone’s GAP representation via element-wise multiplication. Because both streams are pooled to vectors, the difference between \( (H, W) \) and \( (H', W') \) does not affect the fusion. This design preserves pretrained backbone weights and introduces only a lightweight projection layer as additional learnable overhead, making NFP broadly applicable across different architectures.

\begin{figure*}[htb]
\centering

\subfloat[\parbox{2.6cm}{\centering GAP\\ $0.5160$}]{
  \includegraphics[width=0.15\linewidth]{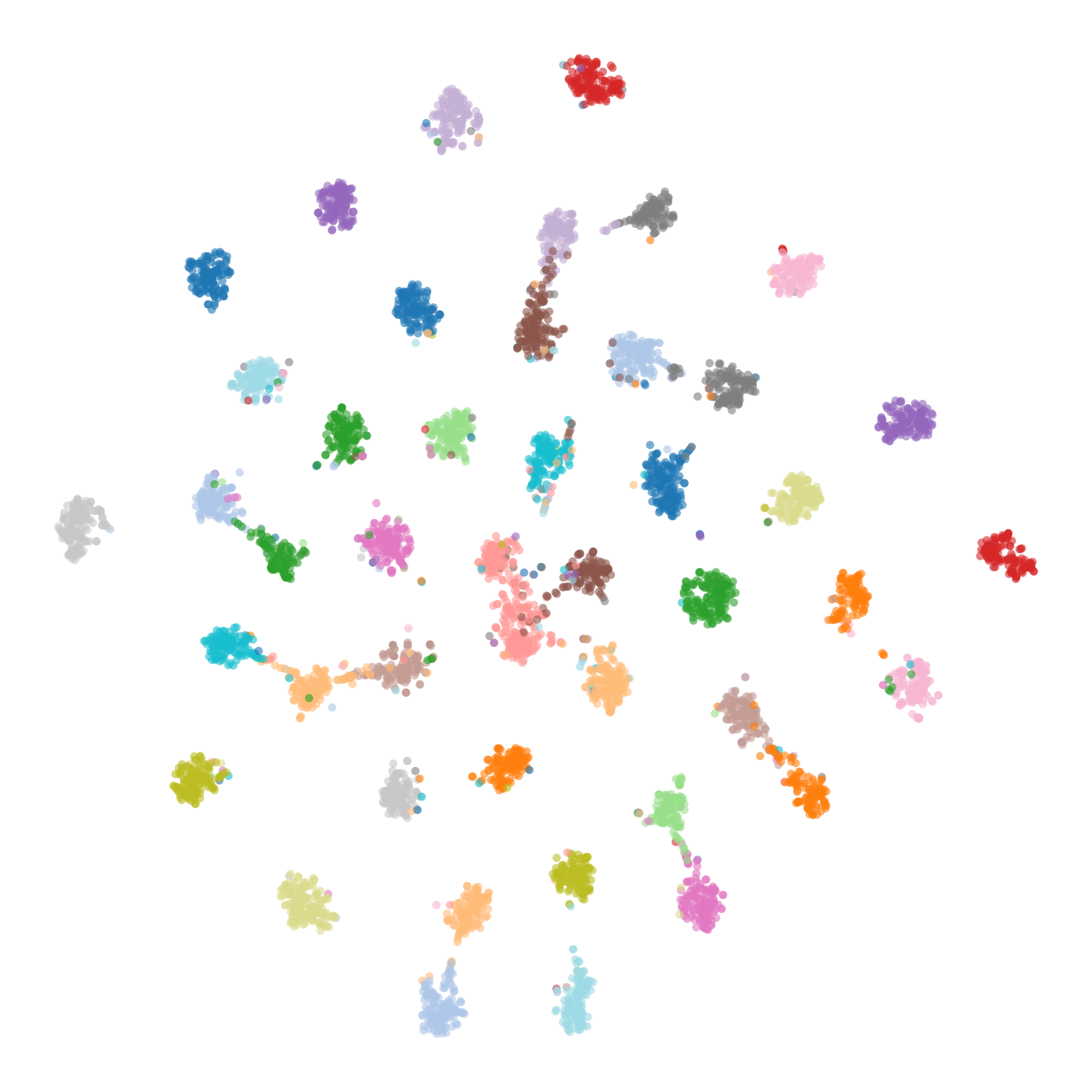}
}
\hfill
\subfloat[\parbox{2.6cm}{\centering Lacunarity\\ $0.4343$}]{
  \includegraphics[width=0.15\linewidth]{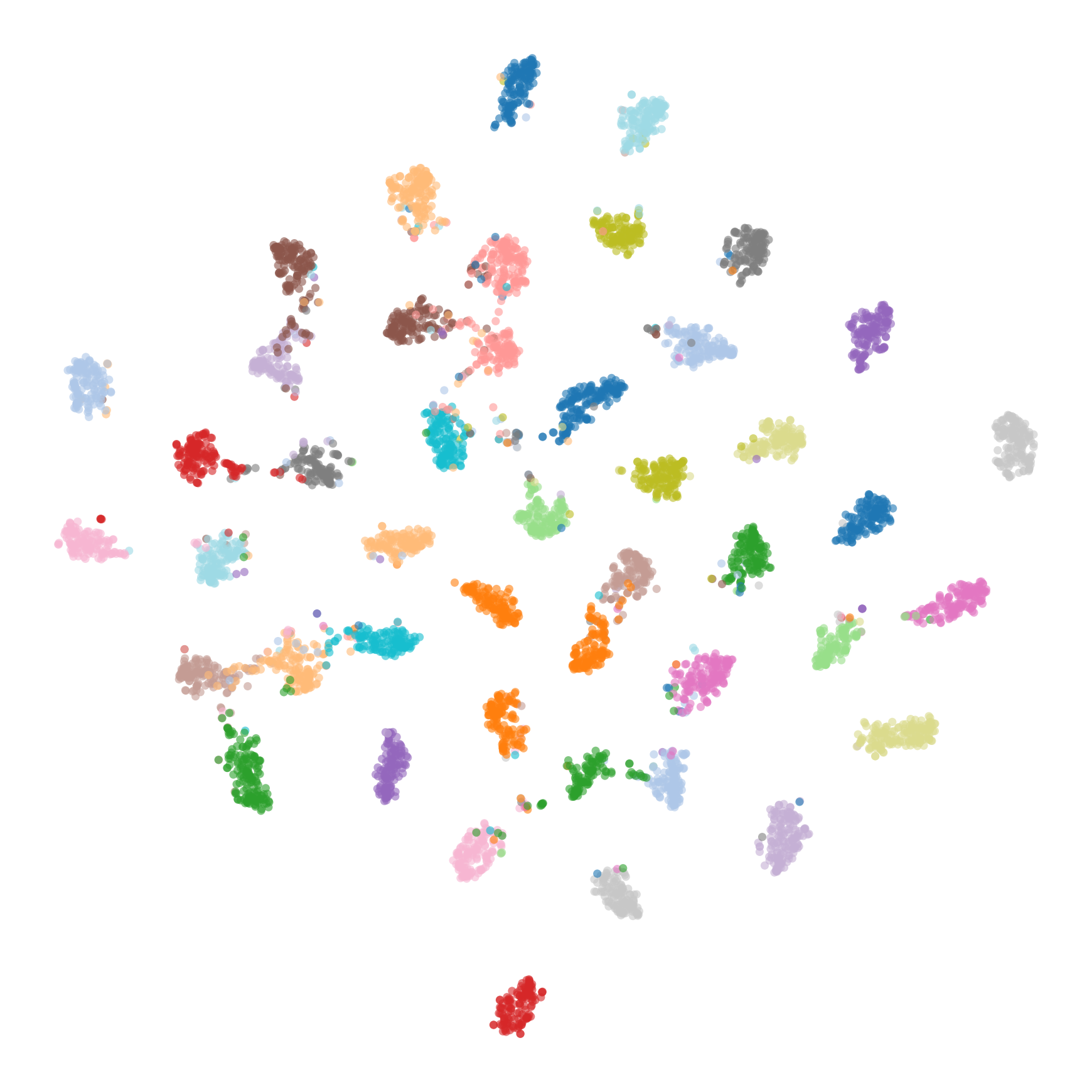}
}
\hfill
\subfloat[\parbox{2.6cm}{\centering Fractal\\ $0.3136$}]{
  \includegraphics[width=0.15\linewidth]{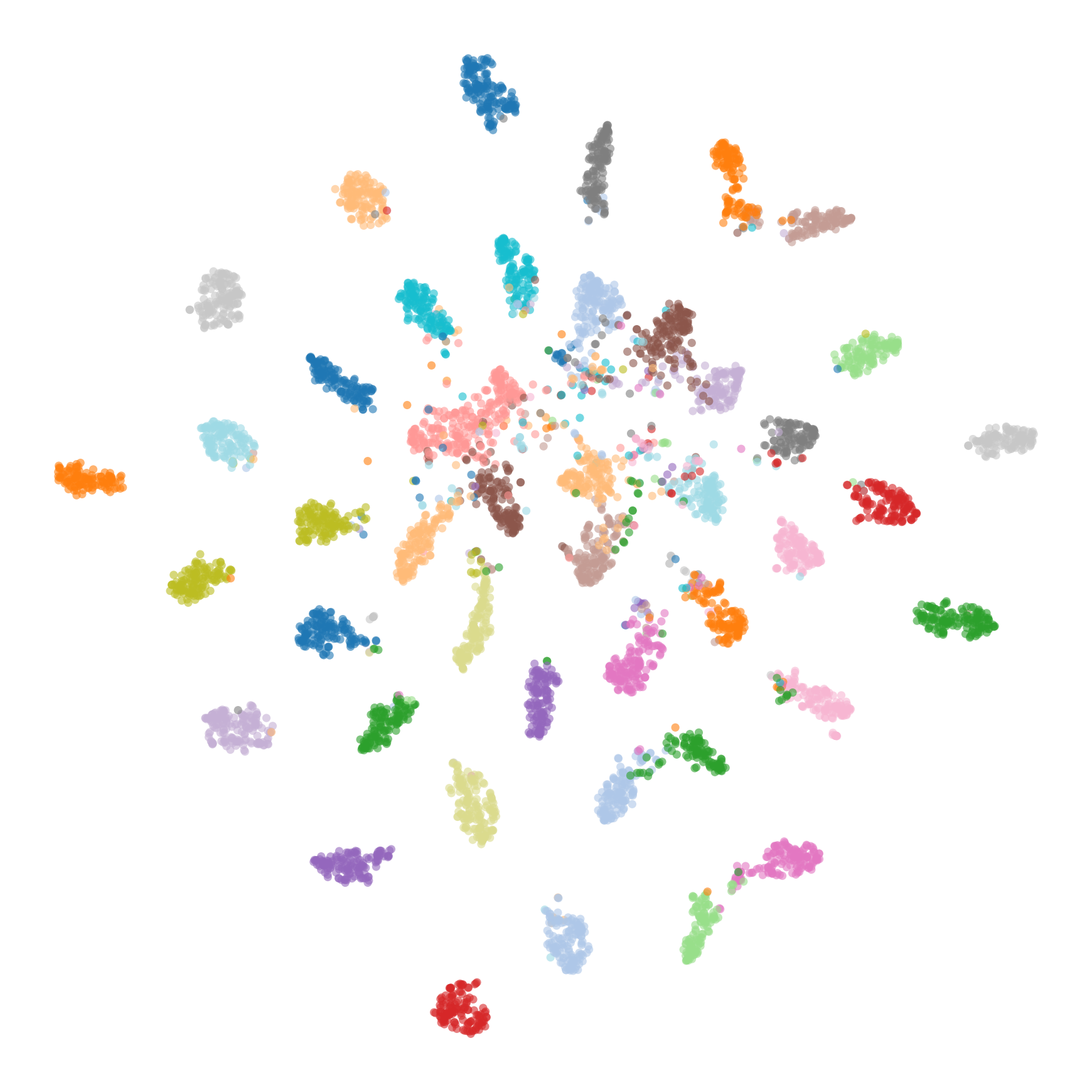}
}
\hfill
\subfloat[\parbox{2.6cm}{\centering RADAM\\ $0.3350$}]{
  \includegraphics[width=0.15\linewidth]{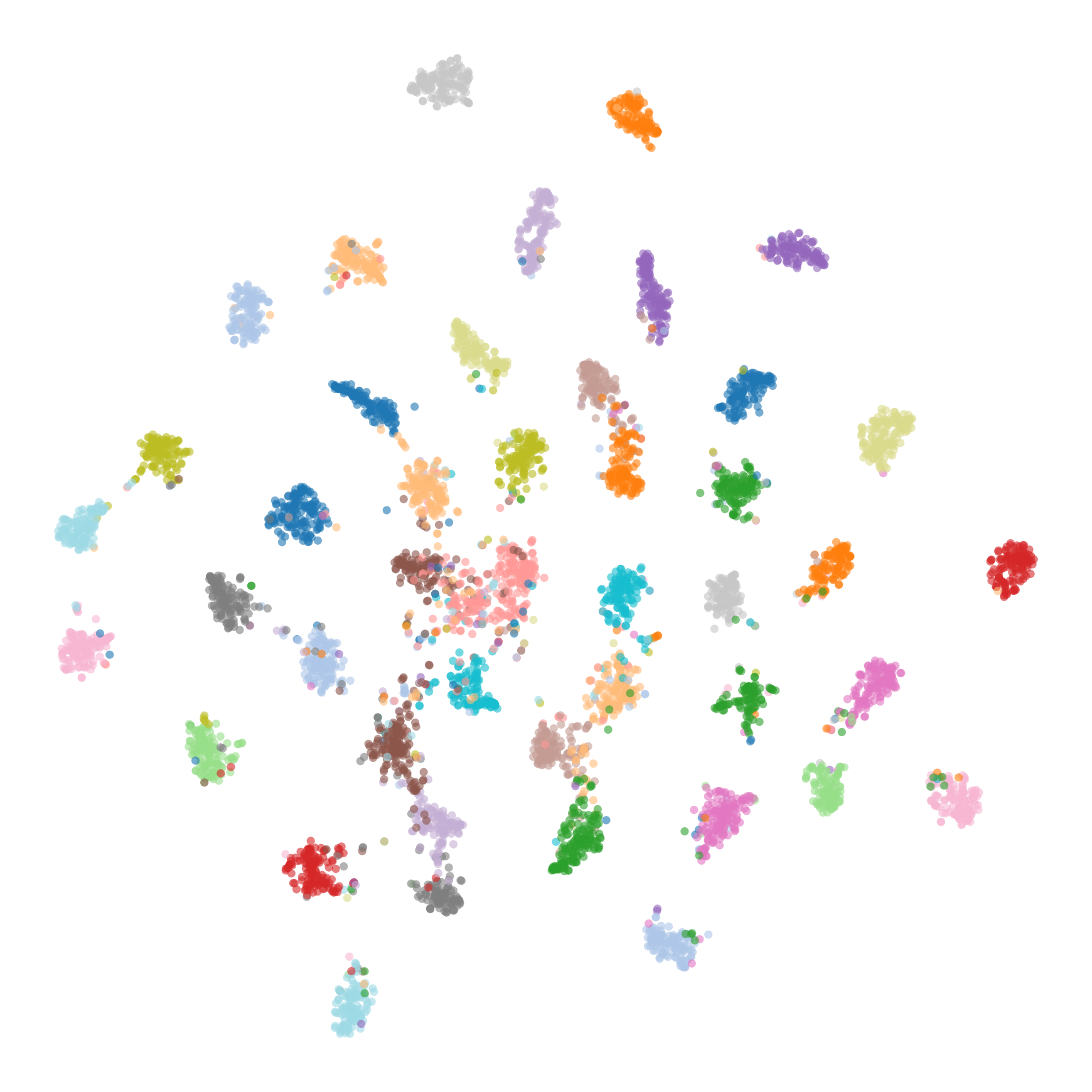}
}
\hfill
\subfloat[\parbox{2.6cm}{\centering DeepTEN\\ $0.5226$}]{
  \includegraphics[width=0.15\linewidth]{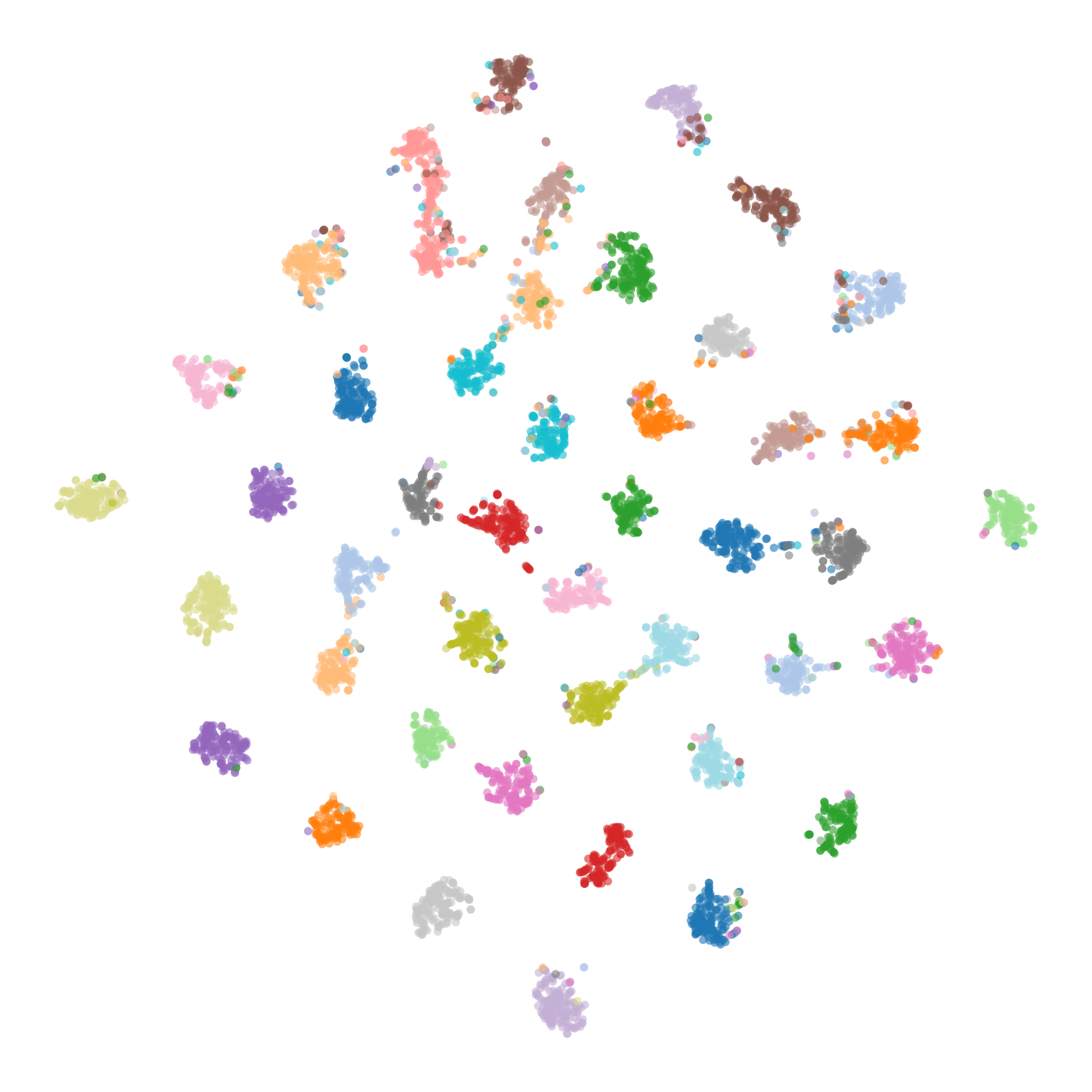}
}
\hfill
\subfloat[\parbox{2.6cm}{\centering NFP (Ours)\\ \textbf{0.5600}}]{
  \includegraphics[width=0.15\linewidth]{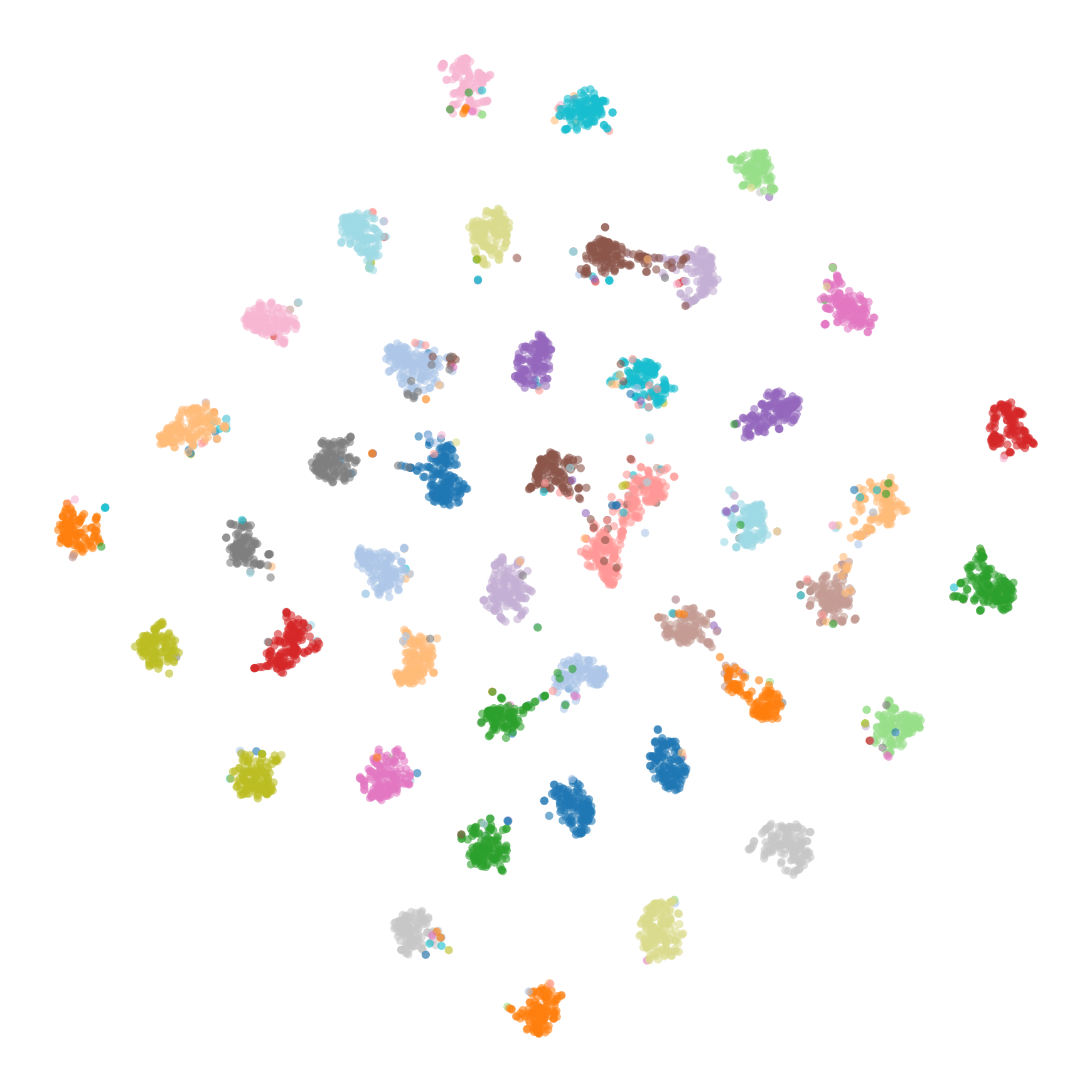}
}

\caption{
t-SNE visualizations of penultimate-layer features extracted from models trained on RESISC45 (45 scene classes) using different pooling methods: GAP, NFP (Ours), DeepTEN, Fractal, Lacunarity, and RADAM. Each point represents a test image, colored by its ground-truth class label. The Silhouette Score, computed from the original feature space, is reported beneath each panel and reflects the compactness and separability of the classes. Among the compared approaches, NFP achieves the highest Silhouette Score, indicating its superior ability to generate more discriminative and well-separated feature embeddings for remote sensing scene classification. All visualizations were generated using a shared random seed to ensure a fair and consistent comparison across methods.}
\label{fig:tsne}
\end{figure*}

\section{Experimental Results and Discussion}
\label{sec:experiment}
\subsection{Experimental setup}

Three public remote sensing datasets are used for evaluation: UC Merced Land Use~\cite{Yang2010UCMerced}, RESISC45~\cite{Cheng2017RESISC45}, and EuroSAT~\cite{helber2019eurosat}. All experiments follow the standard train/validation/test protocols described by Neumann et al.~\cite{neumann2019domain}. To ensure fairness across baselines, all models are trained using identical data splits and preprocessing pipelines. Since EuroSAT contains 13 spectral bands, the ImageNet-initialized first layer is adapted using mean channel-wise initialization, which shows the most consistent performance among tested strategies in~\cref{sec:initialization}.

Experiments are run on a single NVIDIA A100 GPU using three random seeds to report mean and standard deviation. 
Training is performed for up to 100 epochs using the Adam optimizer with a learning rate of 0.001, a batch size of 32, 
and early stopping with a patience of 10 epochs. All images are resized to 256$\times$256. Training-time augmentations 
include horizontal and vertical flips, random rotations ($\pm 15^\circ$), color jitter, and normalization (ImageNet statistics for RGB datasets and per-band statistics for multi-spectral EuroSAT, following~\cite{gomez2021msmatch}). At test time, only resizing and the same normalization are applied. 

\subsection{Backbone Networks}

The NFP layer is evaluated across three convolutional backbone architectures: ConvNeXt-Atto~\cite{liu2022convnet}, MobileNetV3-Large~\cite{howard2019searching}, and ResNet-18~\cite{RESNETDBLP:journals/corr/abs-2110-00476}. All models are initialized with pre-trained weights from the PyTorch Image Models (\texttt{timm}) library~\cite{rw2019timm} and fine-tuned on each dataset. The NFP module is integrated into each backbone according to the scheme described in \cref{subsec:model-arch}. In all cases, NFP is placed after the final convolutional feature extraction stage, followed by a linear classification head. A $1{\times}1$ convolution is applied after NFP to align the number of feature maps to those produced by GAP. Classification is performed by element-wise multiplication of the NFP and GAP feature vectors, followed by a linear classifier. The model architectures for each backbone and pooling variant are trained and evaluated as described above, and results are summarized in \cref{tab:final_grouped_accuracy}.

\subsection{Overall Comparison}
\label{sec:overall_comparison}

\Cref{tab:final_grouped_accuracy} compares the proposed NFP against GAP and recent texture pooling methods (Lacunarity~\cite{mohan2024lacunarity}, Fractal~\cite{florindo2024fractal}, RADAM~\cite{scabini2023radam}, and DeepTEN~\cite{zhang2017deep}) across three remote-sensing datasets and three backbones (ResNet18, MobileNetV3-Large, and ConvNeXt-Atto). Although NFP is conceptually related to neighborhood-similarity modules such as LSP and NSFS and also relies on a cosine-based similarity operator, those modules were originally designed for stereo matching and bi-temporal change detection and are tightly coupled to task-specific architectures. Adapting them into lightweight pooling heads for single-image scene classification would require substantial redesign, so here we focus on controlled comparisons against established texture pooling heads that can be used as drop-in replacements on standard backbones. Accordingly, NFP is best viewed as a lightweight pooling reformulation of neighborhood similarity rather than a new backbone or feature extraction paradigm.

Across all backbone--dataset combinations, NFP is competitive with the strongest baseline and achieves the best or statistically tied accuracy in most cases, while keeping the parameter count essentially identical to the GAP baseline. In the remaining settings, NFP trails the top method by only a very small margin. On simpler benchmarks such as UC Merced, GAP remains a strong baseline and nearly all methods approach saturated performance, but texture-aware pooling still offers a more explicit modeling of neighborhood-level relationships that are attenuated by global average pooling. Methods inspired by handcrafted descriptors such as Lacunarity and Fractal pooling provide modest or inconsistent gains and are constrained by fixed statistical formulations, whereas NFP and other learnable similarity-based pooling heads adapt to the data and yield more reliable improvements across settings.

RADAM, which aggregates activation maps stochastically, can outperform NFP in isolated configurations but exhibits noticeably higher variance across runs, especially on the more challenging dataset–backbone pairs. DeepTEN attains slightly higher accuracy than NFP in a few cases, but these gains are confined to sub-percent differences and come at a substantially higher parameter cost, since DeepTEN consistently enlarges the model relative to the baseline, whereas NFP preserves the lightweight footprint of GAP. This trade-off highlights NFP as an attractive middle ground that captures texture-aware information with minimal complexity and stable optimization behaviour.

\textbf{Backbone-specific trends} For ResNet18, NFP improves over GAP and handcrafted pooling on all three datasets without increasing model size. With MobileNetV3, NFP matches the strongest texture baseline on the easiest dataset and remains very close to the best method on the others, again with negligible parameter overhead. For ConvNeXt-Atto, NFP yields the best performance on two of the three datasets and remains within a narrow margin of the top method on the remaining one, offering a favorable balance between accuracy and efficiency for compact CNN-style backbones.

\textbf{t-SNE Visualization Analysis}
Penultimate-layer features from each model are visualized using t-SNE~\cite{van2008visualizing} on the RESISC45 dataset. \Cref{fig:tsne} presents two-dimensional projections for GAP, Fractal, Lacunarity, RADAM, DeepTEN, and the proposed NFP, with each point colored according to its ground-truth scene class. All visualizations are generated using identical t-SNE hyperparameters and a shared random seed to ensure a fair comparison. The corresponding Silhouette Scores~\cite{rousseeuw1987silhouettes} are reported beneath each plot to quantify class compactness and separability. NFP achieves the highest score (0.5600), followed by GAP (0.5160) and DeepTEN (0.5226). In contrast, Lacunarity (0.4343), RADAM (0.3350), and Fractal (0.3136) exhibit significantly lower values, indicating less class separability and compactness. Both the qualitative t-SNE projections and the quantitative Silhouette Scores demonstrate that NFP produces more compact and better-separated feature embeddings for the RESISC45 dataset.

\begin{table}[htb]
\centering
\begin{tabular}{lccc}
\toprule
Method       & FLOPs & Memory & Latency \\
\midrule
GAP          & 75.80 & 331    & $5.42 \pm 0.83$ \\
Lacunarity   & 75.80 & 549    & $4.89 \pm 0.01$ \\
Fractal      & 76.34 & 593    & $5.21 \pm 0.13$ \\
RADAM        & 75.80 & 636    & $31.94 \pm 0.11$ \\
DeepTEN      & 75.81 & 688    & $6.20 \pm 0.01$ \\
NFP          & 75.89 & 375    & $5.36 \pm 0.01$ \\

\bottomrule
\end{tabular}
\caption{Efficiency comparison of pooling methods on ResNet18 (input size 256$\times$256, batch size 32). FLOPs are reported per forward pass (batch size 32) in gigaflops (G), memory as peak graphics processing unit (GPU) allocation in megabytes (MB), and latency as average per-batch inference time in milliseconds (ms) over multiple forward passes on a single GPU. NFP uses cosine similarity.}
\label{tab:efficiency_comparison}
\end{table}

\begin{figure*}[htb]
\centering
\subfloat[\centering GAP
\\    93.90 ± 0.27 ]{
\includegraphics[width=0.32\linewidth]{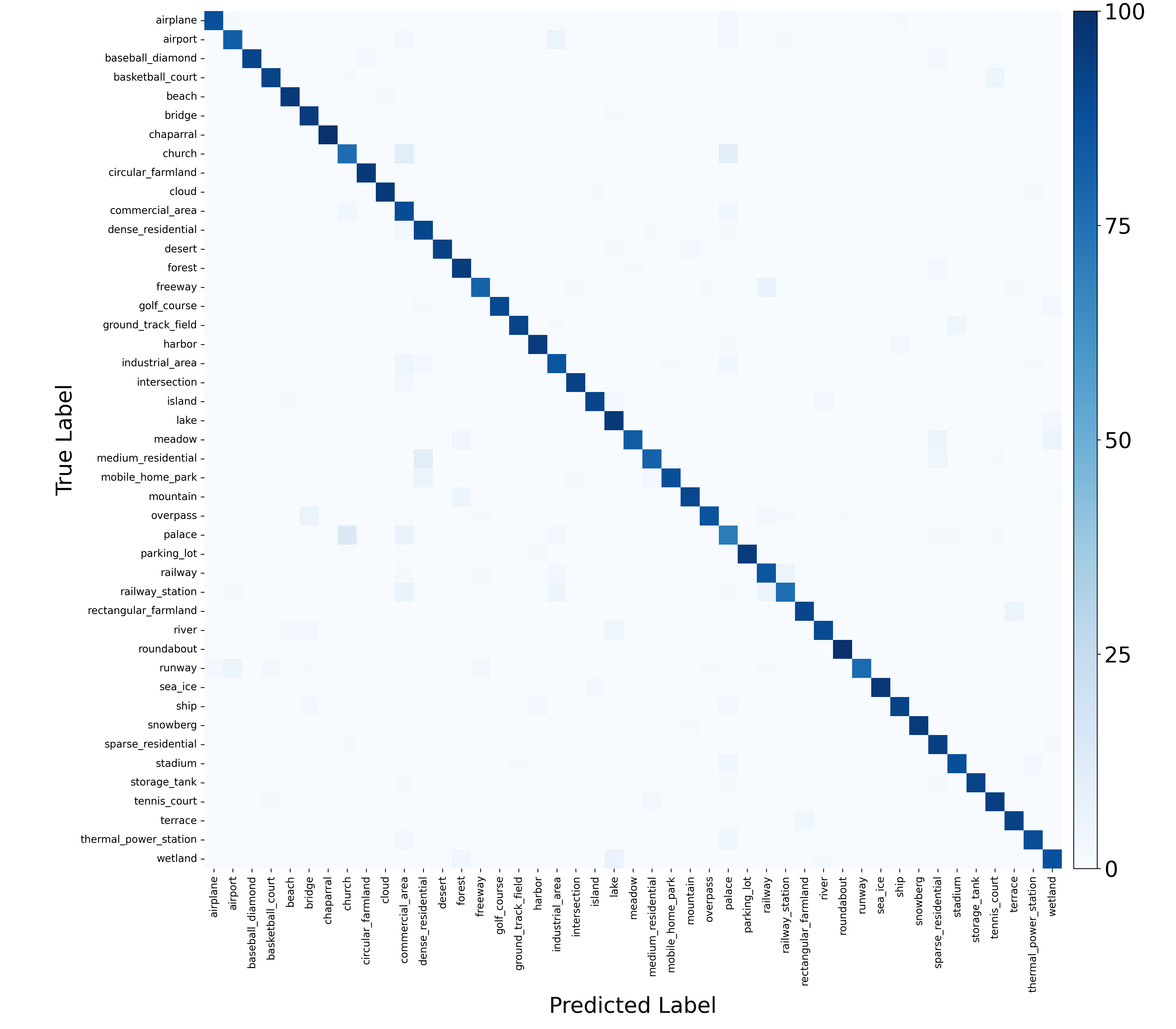}}
\hfill
\subfloat[\centering DeepTEN \\    94.27 ± 0.31]{
\includegraphics[width=0.32\linewidth]{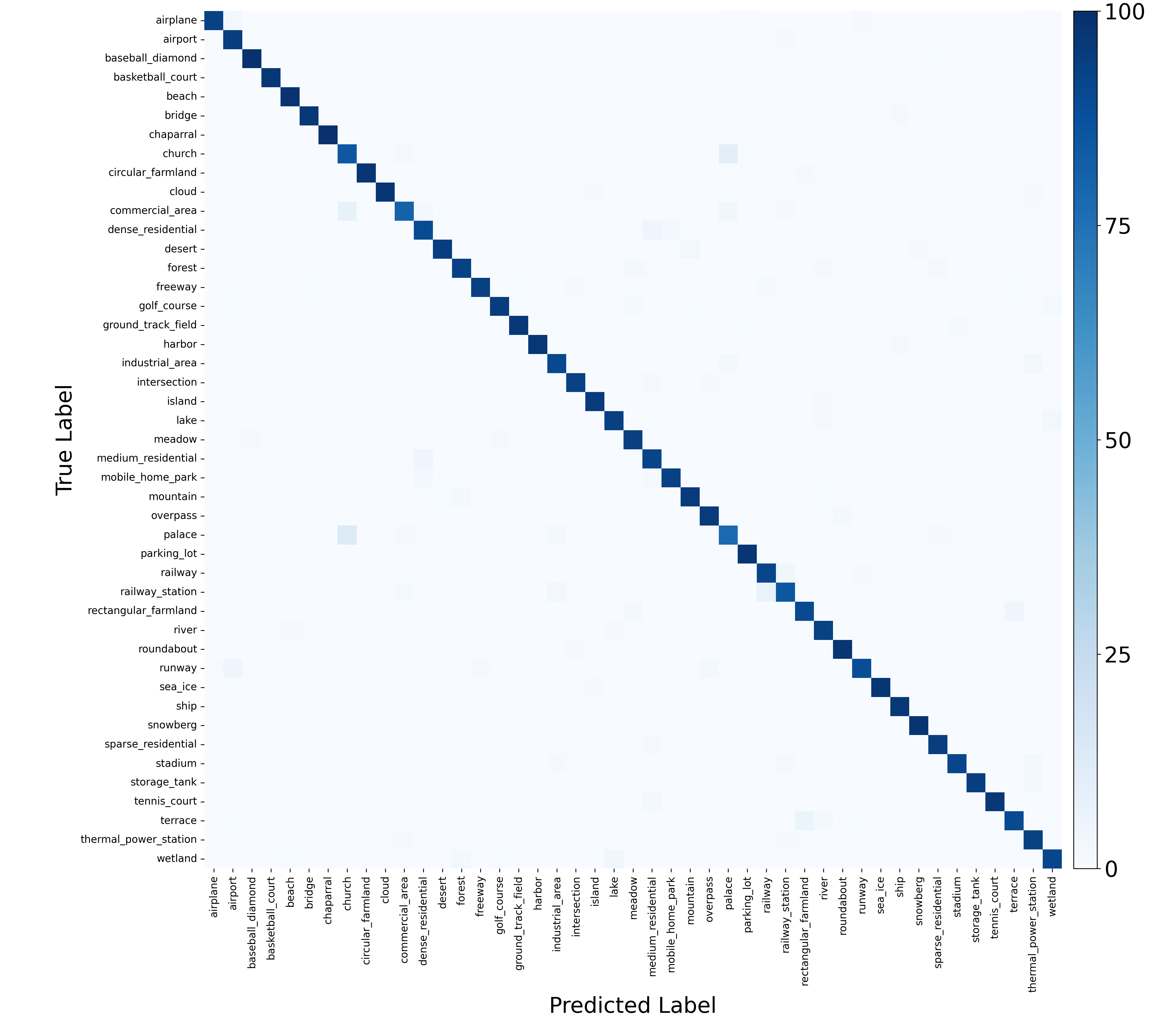}}
\hfill
\subfloat[\centering NFP \\    \textbf{94.71 ± 0.21} ]{
\includegraphics[width=0.32\linewidth]{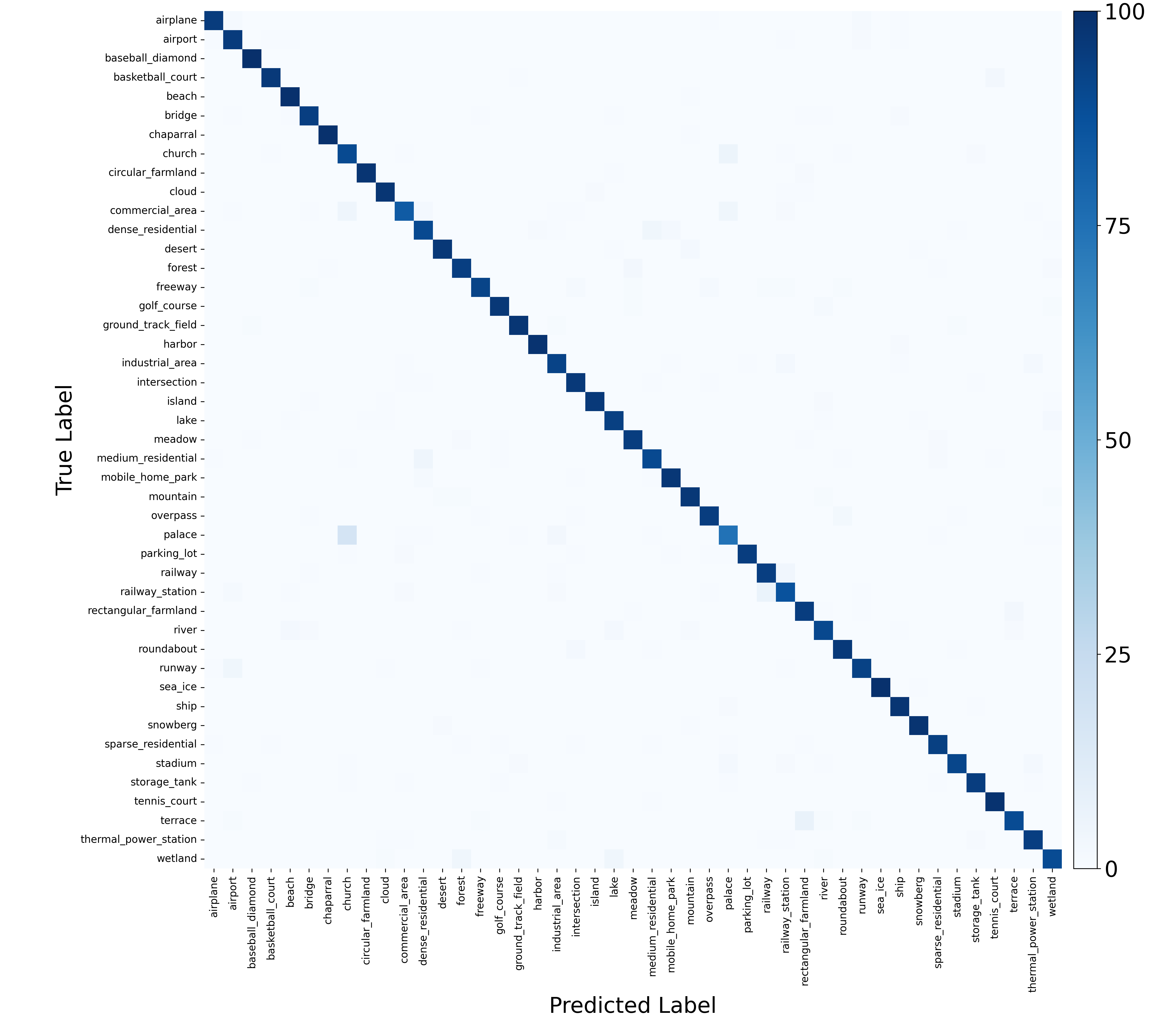}}
\caption{Confusion matrices on the RESISC45 dataset using a ConvNeXt-Atto backbone with three different pooling strategies: GAP, DeepTEN, and the proposed NFP. The NFP variant demonstrates the strongest diagonal dominance and the lowest degree of inter-class confusion.}
\label{fig:confusion_convnext_comparison}
\end{figure*}

\textbf{Computational efficiency} Computational cost is quantified in terms of floating-point operations (FLOPs), peak graphics processing unit (GPU) memory usage, and inference latency for each pooling head combined with a ResNet18 backbone on 256$\times$256 inputs and a batch size of 32. As reported in~\cref{tab:efficiency_comparison}, replacing GAP with NFP (using cosine similarity) yields almost identical FLOPs and latency, with only a modest increase in memory usage. In contrast, Lacunarity, Fractal, DeepTEN, and RADAM exhibit noticeably higher memory requirements, with RADAM in particular showing increased latency. Together with the accuracy gains over GAP in~\cref{tab:final_grouped_accuracy}, these measurements indicate that NFP improves classification performance while largely preserving the computational characteristics of standard GAP.

\textbf{Confusion Matrix Analysis} Figure~\ref{fig:confusion_convnext_comparison} compares the confusion matrices of ConvNeXt backbone using GAP (baseline), DeepTEN, and the proposed NFP on the RESISC45 dataset. The baseline model shows noticeable off-diagonal activations, indicating confusion between visually similar scene categories such as river and lake, terrace and mountain, and dense residential and commercial areas. DeepTEN partially reduces this confusion; however, residual misclassifications remain in several texture-rich and structurally complex classes.

In contrast, the NFP model shows a much cleaner diagonal structure with substantially reduced inter-class confusion. This improvement is most evident in categories with complex spatial layouts and fine-grained textures, including island, river, mobile home park, snowberg, terrace, and wetland. These results highlight the primary advantage of NFP: its ability to capture neighborhood-level similarities and preserve local spatial relationships, leading to stronger class separability and more reliable scene discrimination than both GAP and DeepTEN.

\subsection{Ablation: Similarity Measures}
\label{subsec:similarity-analysis}

We evaluate 18 similarity measures within the NFP framework using ResNet-18 on the EuroSAT dataset to identify the most effective metric for modeling local feature relationships. As shown in~\cref{tab:similarity_comparison}, cosine similarity achieves the best classification performance ($98.52 \pm 0.20$\%), suggesting that angular alignment between local feature vectors is particularly informative for neighborhood encoding. Cosine similarity emphasizes directional agreement while being less sensitive to differences in feature magnitude, which is beneficial in multispectral remote sensing scenes with variations in illumination and contrast. 

Several alternative measures also perform competitively, including Earth Mover’s Distance (EMD) ($97.09 \pm 0.18$\%), Squared Chord ($97.02 \pm 0.17$\%), and L$_2$ norm ($96.96 \pm 0.29$\%). In contrast, performance is notably lower for metrics such as Canberra ($79.14 \pm 1.02$\%) and Chi-Squared Type~2 ($74.35 \pm 7.66$\%), indicating weaker suitability for local neighborhood comparison in this setting. Notably, Sharpened Cosine (SCS) exhibits substantially higher variance ($\pm 10.99$), suggesting unstable performance across runs. Overall, these results support cosine similarity as the default choice for NFP on EuroSAT, providing a strong balance between robustness and discriminative power for local neighborhood modeling.

\subsection{Ablation: Neighborhood Dilation}
\label{subsec:dilation}

To examine the effect of neighborhood sampling distance in NFP, a dilation-based ablation study is conducted on the RESISC45 dataset using a ResNet18 backbone with dilation factors of 1, 3, 5, 7, 9, 11, 13, and 15. Larger neighborhoods could be used to capture broader spatial context at the expense of reduced boundary precision. The neighborhood radius is fixed at \(r = 1\), and only the sampling distance is varied. As reported in \cref{tab:dilation_ablation}, including interpolated intermediate values, performance differences across dilation settings remain minor, indicating that NFP is robust to changes in neighborhood spacing. A dilation of 3 achieves the highest average test accuracy, whereas larger dilation factors do not offer significant differences in performance.

\begin{table}[t]
  \centering
  \begin{tabular}{@{}lc@{}}
    \toprule
    Similarity measure & Accuracy $\pm$ 1 Std (\%) \\
    \midrule
    \textbf{Cosine}                      & \textbf{98.52 $\pm$ 0.20} \\
    Earth Mover's Distance (EMD)        & 97.09 $\pm$ 0.18 \\
    Squared Chord                        & 97.02 $\pm$ 0.17 \\
    L$_2$ Norm                           & 96.96 $\pm$ 0.29 \\
    Smith Similarity                     & 96.96 $\pm$ 0.17 \\
    Chi-Squared Type 1                   & 96.94 $\pm$ 0.05 \\
    Geman-McClure                        & 96.88 $\pm$ 0.20 \\
    Root Mean Square Error (RMSE)        & 96.88 $\pm$ 0.17 \\
    Scaled Dot Product                   & 96.83 $\pm$ 0.24 \\
    GFC                                  & 96.77 $\pm$ 0.12 \\
    Pearson Correlation                  & 96.75 $\pm$ 0.35 \\
    Hellinger                            & 96.68 $\pm$ 0.30 \\
    Jeffrey Divergence                   & 96.23 $\pm$ 0.23 \\
    Sharpened Cosine (SCS)               & 87.67 $\pm$ 10.99 \\
    Canberra                             & 79.14 $\pm$ 1.02 \\
    Chi-Squared Type 2                   & 74.35 $\pm$ 7.66 \\
    L$_1$ Norm                           & 96.96 $\pm$ 0.29 \\
    Dot Product                          & 96.83 $\pm$ 0.24 \\
    \bottomrule
  \end{tabular}
  \caption{
    Comparison of 18 similarity measures for NFP using ResNet-18 on the EuroSAT dataset. Best average result is shown in \textbf{bold}.
  }
  \label{tab:similarity_comparison}
\end{table}

\begin{table}[t]
\centering
\small
\begin{tabular}{cc}
\toprule
Dilation Factor & Accuracy $\pm$ 1 Std (\%) \\
\midrule
1   & 94.02 $\pm$ 0.25 \\
3   & \textbf{94.42 $\pm$ 0.10} \\
5   & 94.16 $\pm$ 0.05 \\
7   & 94.28 $\pm$ 0.07 \\
9   & 94.11 $\pm$ 0.09 \\
11  & 94.22 $\pm$ 0.06 \\
13  & 94.08 $\pm$ 0.11 \\
15  & 94.23 $\pm$ 0.04 \\
\bottomrule
\end{tabular}
\caption{
Effect of neighborhood dilation on NFP performance using ResNet18 on the RESISC45 dataset. Best average result is shown in \textbf{bold}. Intermediate values are interpolated for trend visualization.
}
\label{tab:dilation_ablation}
\end{table}

\subsection{Ablation: Layer Placement}
\label{subsec:layer-replacement}

\begin{table}[htb]
\centering
\setlength{\tabcolsep}{2pt} 
\renewcommand{\arraystretch}{1.1} 
\begin{tabular}{lccc}
\toprule
Network Stage & Channels & Acc. $\pm$ 1 Std (\%) & Params \\
\midrule
Layer~1 & 16   & 98.73 $\pm$ 0.11 & 4.24M \\
Layer~2 & 24   & 98.65 $\pm$ 0.11 & 4.24M \\
Layer~3 & 40   & 98.33 $\pm$ 0.00 & 4.24M \\
Layer~4 & 112  & 98.02 $\pm$ 0.11 & 4.25M \\
Layer~5 & 960  & 98.33 $\pm$ 0.51 & 4.31M \\
All & 1152 & \textbf{98.89 $\pm$ 0.11} & 4.37M \\
\bottomrule
\end{tabular}
\caption{
Classification accuracy, number of channels before NFP (cosine similarity), and number of trainable parameters for different insertion stages of MobileNetV3~\cite{howard2019searching} on UC~Merced~\cite{Yang2010UCMerced}. Channels are from \texttt{features\_only} output~\cite{rw2019timm}. Best average result is shown in \textbf{bold}.
}
\label{tab:early_layer_nfp}
\end{table}

The effect of inserting the NFP layer at different depths of the backbone is further explored to assess whether early-stage spatial patterns can improve texture-aware classification. Specifically, the MobileNetV3~\cite{howard2019searching} architecture is evaluated by placing the NFP module after each of its main feature extraction stages (Layers~1 through~5), as defined by the \texttt{features\_only} outputs of the \texttt{timm} library. In each configuration, features from the selected stage are passed through the NFP module, followed by a classification head.

This experimental setup isolates the contribution of each stage’s feature map, enabling a systematic evaluation of where texture information is most beneficial. As summarized in \cref{tab:early_layer_nfp}, the highest accuracy is achieved when NFP aggregates features from all stages (98.89 $\pm$ 0.11), providing a modest gain over the best single-stage placement. Among individual stages, the strongest result is obtained at Layer 1 (98.73 $\pm$ 0.11), with accuracy gradually declining at deeper stages. These results indicate that early-stage features carry strong discriminative texture cues, while multi-stage aggregation further improves performance with a small parameter increase.

\subsection{Ablation: Initialization Strategy}
\label{sec:initialization}
EuroSAT has 13 spectral bands, so the first layer of the network must be modified to handle more input channels than the standard RGB case. The results in \cref{tab:init_ablation} show that mean channel-wise initialization gives the best accuracy for both GAP and the proposed NFP. This suggests that averaging the pretrained RGB weights provides a better starting point for learning from multispectral data. In contrast, random initialization removes useful pretrained information and leads to worse performance, while weight replication and the $1{\times}1$ adapter are less effective. In addition, the $1{\times}1$ adapter adds extra parameters, whereas mean initialization keeps the model lightweight. For these reasons, mean initialization is used in all other experiments.

\subsection{Ablation: Fusion Strategy}
\label{sec:fusion}
\begin{table}[t]
\centering
\begin{tabular}{lcc}
\toprule
Init. Strategy & GAP (\%) & NFP (\%) \\
\midrule
Weight Replication   & $97.85 \pm 0.25$ & $98.05 \pm 0.22$ \\
\textbf{Mean Init.}  & $\mathbf{98.27 \pm 0.19}$ & $\mathbf{98.52 \pm 0.20}$ \\
$1{\times}1$ Adapter & $97.42 \pm 0.28$ & $97.91 \pm 0.26$ \\
Random Init.         & $94.63 \pm 0.83$ & $94.80 \pm 0.78$ \\
\bottomrule
\end{tabular}
\caption{Effect of different 13-channel initialization (init) strategies on EuroSAT. Each entry reports average accuracy $\pm$ 1 standard deviation across multiple runs for both GAP and NFP. Mean initialization corresponds to the final configuration and yields the highest performance.}
\label{tab:init_ablation}
\end{table}

    The proposed design combines NFP and GAP via element-wise product, allowing local neighborhood similarity responses to rescale global activations. To assess whether more expressive interactions are necessary, three alternative fusion mechanisms were evaluated on UC Merced with a ResNet18 backbone: (i) simple concatenation, (ii) concatenation followed by a two-layer MLP, and (iii) Squeeze-and-Excitation (SE)~\cite{hu2018squeeze} style gated weighting, where a scalar weight $\alpha$ balances GAP and NFP features. As summarized in~\cref{tab:fusion_ablation}, the element-wise product achieves the highest accuracy while using the second fewest parameters. This suggests that the element-wise product preserves complementary information from both NFP and GAP, which creates a more aligned and interpretable interaction between the two branches, while avoiding unnecessary parameter growth. In contrast, the MLP and SE-style gating increase the model size from 11.23M to 12.02M and 12.28M parameters, respectively, without providing a proportional improvement in performance.

\begin{table}[t]
\centering
\small
\setlength{\tabcolsep}{3.5pt}
\begin{tabular}{@{}lcc@{}}
\toprule
\textbf{Fusion type} & \textbf{Params (M)} & \textbf{Acc. $\pm$ Std (\%)} \\
\midrule
Concatenation only  & 11.19 & 96.35 ± 0.15 \\
Concatenation + MLP & 12.02 & 97.05 ± 0.13 \\
Gated weighting (SE-style) & 12.28 & 98.12 ± 0.25 \\
\textbf{Element-wise product (Ours)} & {11.23} & \textbf{98.86 ± 0.58} \\
\bottomrule
\end{tabular}
\caption{Comparison of fusion strategies for combining NFP and GAP on the UC Merced dataset (ResNet18 backbone). Parameter count is given in millions. Best average result is shown in \textbf{bold}.}
\label{tab:fusion_ablation}
\end{table}

\section{Conclusion}
\label{sec:conclusion}

This work introduces NFP, a method designed to enhance texture-aware classification in remote sensing images. By explicitly modeling local similarity relationships within the feature space, NFP enables convolutional backbones to preserve fine-grained spatial structure alongside global semantic information. Extensive experiments across three public remote-sensing datasets and three backbone architectures demonstrate that NFP is competitive with, and often surpasses, state-of-the-art texture pooling methods, including GAP, Lacunarity, Fractal, and RADAM, as well as DeepTEN, while keeping the additional parameter cost minimal and preserving the lightweight footprint of the underlying backbones.

Further analysis of similarity metrics, layer placement, neighborhood dilation, and visualizations (\eg, t-SNE) confirm that NFP produces more discriminative and semantically meaningful feature embeddings compared to prior methods. A moderate dilation factor was found to slightly improve performance without increasing parameters, indicating that limited expansion of spatial context can be beneficial. These benefits are most pronounced on compact CNN backbones such as ResNet18, MobileNetV3, and ConvNeXt-Atto. Future work includes exploring learnable weighting schemes for combining GAP and NFP, more advanced multi-scale or adaptive neighborhood formulations (e.g., radial windows), as well as extending NFP toward domain adaptation and lightweight deployment in diverse vision tasks (e.g., object detection and segmentation).

\section*{Acknowledgment}
Portions of this research were conducted with the advanced computing resources provided by Texas A\&M High Performance Research Computing. LA-UR-2530363.

{
    \small
    \bibliographystyle{ieeenat_fullname}
    \bibliography{main}

@article{avi2023differentiable,
  title={Differentiable histogram loss functions for intensity-based image-to-image translation},
  author={Avi-Aharon, Mor and Arbelle, Assaf and Raviv, Tammy Riklin},
  journal={IEEE Transactions on Pattern Analysis and Machine Intelligence},
  volume={45},
  number={10},
  pages={11642--11653},
  year={2023},
  publisher={IEEE}
}

@article{puissant2005utility,
  title={The utility of texture analysis to improve per-pixel classification for high to very high spatial resolution imagery},
  author={Puissant, Anne and Hirsch, Jacky and Weber, Christiane},
  journal={International Journal of Remote Sensing},
  volume={26},
  number={4},
  pages={733--745},
  year={2005},
  publisher={Taylor \& Francis}
}

@article{peeples2021histogram,
  title={Histogram layers for texture analysis},
  author={Peeples, Joshua and Xu, Weihuang and Zare, Alina},
  journal={IEEE Transactions on Artificial Intelligence},
  volume={3},
  number={4},
  pages={541--552},
  year={2021},
  publisher={IEEE}
}

@article{scabini2023radam,
  title   = {RADAM: Texture recognition through randomized aggregated encoding of deep activation maps},
  author  = {Scabini, Leonardo and Zielinski, Kallil M and Ribas, Lucas C and Gon{\c{c}}alves, Wesley N and De Baets, Bernard and Bruno, Odemir M},
  journal = {Pattern Recognition},
  volume  = {143},
  pages   = {109802},
  year    = {2023},
  publisher = {Elsevier}
}

@article{florindo2024fractal,
  title   = {Fractal pooling: A new strategy for texture recognition using convolutional neural networks},
  author  = {Florindo, Jo{\~a}o B},
  journal = {Expert Systems with Applications},
  volume  = {243},
  pages   = {122978},
  year    = {2024},
  publisher = {Elsevier}
}

@inproceedings{mohan2024lacunarity,
  author    = {Mohan, Akshatha and Peeples, Joshua},
  title     = {Lacunarity Pooling Layers for Plant Image Classification using Texture Analysis},
  booktitle = {Proceedings of the IEEE/CVF Conference on Computer Vision and Pattern Recognition (CVPR) Workshops},
  month     = {June},
  year      = {2024},
  pages     = {5384--5392}
}

@article{ahmed2011compound,
  title={Compound local binary pattern (clbp) for rotation invariant texture classification},
  author={Ahmed, Faisal and Hossain, Emam and Bari, ASMH and Hossen, Md Sakhawat},
  journal={International Journal of Computer Applications},
  volume={33},
  number={6},
  pages={5--10},
  year={2011},
  publisher={International Journal of Computer Applications, 244 5 th Avenue,\# 1526, New~…}
}

@article{ojala2002multiresolution,
  title={Multiresolution gray-scale and rotation invariant texture classification with local binary patterns},
  author={Ojala, Timo and Pietikainen, Matti and Maenpaa, Topi},
  journal={IEEE Transactions on pattern analysis and machine intelligence},
  volume={24},
  number={7},
  pages={971--987},
  year={2002},
  publisher={IEEE}
}

@inproceedings{zhang2017deep,
  title     = {DeepTEN: Texture Encoding Network},
  author    = {Zhang, Hang and Xue, Jia and Dana, Kristin},
  booktitle = {Proceedings of the IEEE Conference on Computer Vision and Pattern Recognition (CVPR)},
  pages     = {708--717},
  year      = {2017}
}

@inproceedings{Liu2022LSP,
  title     = {Local Similarity Pattern and Cost Self-Reassembling for Deep Stereo Matching Networks},
  author    = {Liu, Biyang and Yu, Huimin and Long, Yangqi},
  booktitle = {Proceedings of the AAAI Conference on Artificial Intelligence},
  year      = {2022},
  pages     = {1647--1654}
}

@inproceedings{HampelArias2023FeatureLayers,
  title     = {Experiments in Anomalous Change Detection: Improving Detector Discrimination through Feature Layers},
  author    = {Hampel-Arias, Zigfried and Ziemann, Amanda},
  booktitle = {Algorithms, Technologies, and Applications for Multispectral and Hyperspectral Imaging XXIX, Proc.\ SPIE},
  volume    = {12519},
  pages     = {125190P},
  year      = {2023},
  doi       = {10.1117/12.2666569}
}

@ARTICLE{7061924,
  author  = {Deborah, Hilda and Richard, No{\"e}l and Hardeberg, Jon Yngve},
  journal = {IEEE Journal of Selected Topics in Applied Earth Observations and Remote Sensing},
  title   = {A Comprehensive Evaluation of Spectral Distance Functions and Metrics for Hyperspectral Image Processing},
  year    = {2015},
  volume  = {8},
  number  = {6},
  pages   = {3224--3234},
  doi     = {10.1109/JSTARS.2015.2403257}
}

@article{wu2023exploring,
  title   = {Exploring the sharpened cosine similarity},
  author  = {Wu, Skyler and Lu, Fred and Raff, Edward and Holt, James},
  journal = {arXiv preprint arXiv:2307.13855},
  year    = {2023}
}

@inproceedings{howard2019searching,
  title     = {Searching for MobileNetV3},
  author    = {Howard, Andrew and Sandler, Mark and Chu, Grace and Chen, Liang-Chieh and Chen, Bo and Tan, Mingxing and Wang, Weijun and Zhu, Yukun and Pang, Ruoming and Vasudevan, Vijay and others},
  booktitle = {Proceedings of the IEEE/CVF International Conference on Computer Vision (ICCV)},
  pages     = {1314--1324},
  year      = {2019}
}

@article{RESNETDBLP:journals/corr/abs-2110-00476,
  author       = {Wightman, Ross and Touvron, Hugo and J{\'e}gou, Herv{\'e}},
  title        = {ResNet strikes back: An improved training procedure in timm},
  journal      = {CoRR},
  volume       = {abs/2110.00476},
  year         = {2021},
  url          = {https://arxiv.org/abs/2110.00476},
  eprinttype   = {arXiv},
  eprint       = {2110.00476}
}

@misc{rw2019timm,
  author       = {Wightman, Ross},
  title        = {PyTorch Image Models},
  year         = {2019},
  howpublished = {\url{https://github.com/rwightman/pytorch-image-models}},
  doi          = {10.5281/zenodo.4414861},
  note         = {Accessed: 2025-08-17}
}

@inproceedings{Yang2010UCMerced,
  author    = {Yang, Yi and Newsam, Shawn D.},
  title     = {Bag-of-Visual-Words and Spatial Extensions for Land-Use Classification},
  booktitle = {Proceedings of the 18th ACM SIGSPATIAL International Conference on Advances in Geographic Information Systems (GIS)},
  year      = {2010},
  pages     = {270--279},
  doi       = {10.1145/1869790.1869829}
}

@article{Cheng2017RESISC45,
  author  = {Cheng, Gong and Han, Junwei and Lu, Xiaoqiang},
  title   = {Remote Sensing Image Scene Classification: Benchmark and State of the Art},
  journal = {Proceedings of the IEEE},
  volume  = {105},
  number  = {10},
  pages   = {1865--1883},
  year    = {2017},
  doi     = {10.1109/JPROC.2017.2675998}
}

@article{helber2019eurosat,
  title   = {EuroSAT: A Novel Dataset and Deep Learning Benchmark for Land Use and Land Cover Classification},
  author  = {Helber, Patrick and Bischke, Benjamin and Dengel, Andreas and Borth, Damian},
  journal = {IEEE Journal of Selected Topics in Applied Earth Observations and Remote Sensing},
  volume  = {12},
  number  = {7},
  pages   = {2217--2226},
  year    = {2019},
  publisher = {IEEE}
}

@article{gomez2021msmatch,
  title   = {MSMatch: Semisupervised Multispectral Scene Classification with Few Labels},
  author  = {G{\'o}mez, Pablo and Meoni, Gabriele},
  journal = {IEEE Journal of Selected Topics in Applied Earth Observations and Remote Sensing},
  volume  = {14},
  pages   = {11643--11654},
  year    = {2021},
  publisher = {IEEE}
}

@article{neumann2019domain,
  title   = {In-domain Representation Learning for Remote Sensing},
  author  = {Neumann, Maxim and Pinto, Andr{\'e} Susano and Zhai, Xiaohua and Houlsby, Neil},
  journal = {arXiv preprint arXiv:1911.06721},
  year    = {2019}
}

@article{van2008visualizing,
  title   = {Visualizing Data Using t-SNE},
  author  = {Van der Maaten, Laurens and Hinton, Geoffrey},
  journal = {Journal of Machine Learning Research},
  volume  = {9},
  number  = {Nov},
  pages   = {2579--2605},
  year    = {2008}
}

@article{rousseeuw1987silhouettes,
  title   = {Silhouettes: A Graphical Aid to the Interpretation and Validation of Cluster Analysis},
  author  = {Rousseeuw, Peter J.},
  journal = {Journal of Computational and Applied Mathematics},
  volume  = {20},
  pages   = {53--65},
  year    = {1987},
  publisher = {Elsevier}
}

@article{liu2022convnet,
  author  = {Zhuang Liu and Hanzi Mao and Chao-Yuan Wu and Christoph Feichtenhofer and Trevor Darrell and Saining Xie},
  title   = {A ConvNet for the 2020s},
  journal = {Proceedings of the IEEE/CVF Conference on Computer Vision and Pattern Recognition (CVPR)},
  year    = {2022},
}

@article{Kerr2003,
  author  = {Kerr, J. T. and Ostrovsky, M.},
  title   = {From Space to Species: Ecological Applications of Remote Sensing},
  journal = {Trends in Ecology \& Evolution},
  volume  = {18},
  number  = {6},
  pages   = {299--305},
  year    = {2003}
}

@article{Wellmann2020Urban,
  author  = {Wellmann, Thilo and Lausch, Angela and Andersson, Erik and others},
  title   = {Remote Sensing in Urban Planning: Contributions towards Ecologically Sound Policies?},
  journal = {Landscape and Urban Planning},
  volume  = {204},
  pages   = {103921},
  year    = {2020},
  doi     = {10.1016/j.landurbplan.2020.103921}
}

@article{Mulla2013,
  author  = {Mulla, David J.},
  title   = {Twenty Five Years of Remote Sensing in Precision Agriculture: Key Advances and Remaining Knowledge Gaps},
  journal = {Biosystems Engineering},
  volume  = {114},
  number  = {4},
  pages   = {358--371},
  year    = {2013},
  doi     = {10.1016/j.biosystemseng.2012.08.009}
}

@article{Zhang2016Tutorial,
  author  = {Zhang, Liangpei and Zhang, Lefei and Du, Bo},
  title   = {Deep Learning for Remote Sensing Data: A Technical Tutorial on the State of the Art},
  journal = {IEEE Geoscience and Remote Sensing Magazine},
  volume  = {4},
  number  = {2},
  pages   = {22--40},
  year    = {2016},
  doi     = {10.1109/MGRS.2016.2540798}
}

@inproceedings{hu2018squeeze,
  title={Squeeze-and-excitation networks},
  author={Hu, Jie and Shen, Li and Sun, Gang},
  booktitle={Proceedings of the IEEE conference on computer vision and pattern recognition},
  pages={7132--7141},
  year={2018}
}

@article{Li2018Survey,
  author  = {Li, Ying and Zhang, Haokui and Xue, Xizhe and Jiang, Yenan and Shen, Qiang},
  title   = {Deep Learning for Remote Sensing Image Classification: A Survey},
  journal = {Wiley Interdisciplinary Reviews: Data Mining and Knowledge Discovery},
  volume  = {8},
  number  = {6},
  pages   = {e1264},
  year    = {2018},
  doi     = {10.1002/widm.1264}
}

@article{zhao2024improved,
  title={A improved pooling method for convolutional neural networks},
  author={Zhao, Lei and Zhang, Zhonglin},
  journal={Scientific Reports},
  volume={14},
  number={1},
  pages={1589},
  year={2024},
  publisher={Nature Publishing Group UK London}
}

@inproceedings{ojala1994performance,
  title     = {Performance Evaluation of Texture Measures with Classification Based on Kullback Discrimination of Distributions},
  author    = {Ojala, Timo and Pietik{\"a}inen, Matti and Harwood, David},
  booktitle = {Proceedings of the 12th International Conference on Pattern Recognition (ICPR)},
  volume    = {1},
  pages     = {582--585},
  year      = {1994},
  organization = {IEEE}
}
}

\balance
\end{document}